\def\cvprPaperID{12361}
\def\confName{ICCV}
\def\confYear{2023}

\def\paperTitle{Generalizable Decision Boundaries: Dualistic Meta-Learning for \\ Open Set Domain Generalization}

\def\authorBlock{
    Xiran Wang$^{1}$ \qquad
    Jian Zhang$^{1}$ \qquad
    Lei Qi$^{2}$ \qquad
    Yinghuan Shi$^{1}$\thanks{
    Corresponding author: Yinghuan Shi. 
    Xiran Wang, Jian Zhang and Yinghuan Shi are with the State Key Laboratory for Novel Software Technology and National Institute of Healthcare Data Science, Nanjing University, China. 
    Lei Qi is with the School of Computer Science and Engineering, Southeast University, China. 
    This work is supported by NSFC Program (62222604, 62206052, 62192783), China Postdoctoral Science Foundation Project (2023T160100), Jiangsu Natural Science Foundation Project (BK20210224), and CCF-Lenovo Bule Ocean Research Fund.} \\
    $^{1}$Nanjing University \qquad
    $^{2}$Southeast University \\
    {\tt\small \{zzwdx, zhangjian7369\}@smail.nju.edu.cn, qilei@seu.edu.cn, syh@nju.edu.cn}
}

\newif\ifreview 
\newif\ifarxiv 
\newif\ifcamera \newcommand{\cameraready}{\cameratrue}
\newif\ifrebuttal 

\cameraready 

\pdfoutput=1
\documentclass[10pt,twocolumn,letterpaper]{article}
\ifreview \usepackage[review]{cvpr} \fi
\ifarxiv \usepackage[pagenumbers]{cvpr} \fi
\ifrebuttal \usepackage[rebuttal]{cvpr} \fi
\ifcamera \usepackage{cvpr} \fi

\usepackage{graphicx}
\usepackage{amsmath}
\usepackage{amssymb}
\usepackage{booktabs}

\usepackage{times}
\usepackage{microtype}
\usepackage{epsfig}
\usepackage[table,xcdraw]{xcolor}
\usepackage{caption}
\usepackage{float}
\usepackage{placeins}
\usepackage{color, colortbl}
\usepackage{stfloats}
\usepackage{enumitem}
\usepackage{tabularx}
\usepackage{xstring}
\usepackage{multirow}
\usepackage{xspace}
\usepackage{url}
\usepackage{subcaption}
\usepackage{xcolor}
\usepackage{bbding}
\usepackage{makecell}
\usepackage{algorithm} 
\usepackage{algorithmic}
\usepackage{wrapfig}

\ifcamera \usepackage[accsupp]{axessibility} \fi

% ...

\ifarxiv  \fi

\newcommand{\R}[1]{{%
    \textbf{%
        \ifstrequal{#1}{1}{\textcolor{red}{R#1}}{%
        \ifstrequal{#1}{2}{\textcolor{blue}{R#1}}{%
        \ifstrequal{#1}{3}{\textcolor{magenta}{R#1}}{%
        \ifstrequal{#1}{4}{\textcolor{teal}{R#1}}{%
                           \textcolor{cyan}{R#1}%
        }}}}%
    }%
}}

\usepackage{xr-hyper}

\makeatletter
\newcommand*{\addFileDependency}[1]{
  \typeout{(#1)}
  \@addtofilelist{#1}
  \IfFileExists{#1}{}{\typeout{No file #1.}}
}

\makeatother

\usepackage[pagebackref,breaklinks,colorlinks]{hyperref}
\usepackage[capitalize]{cleveref}
\crefname{section}{Sec.}{Secs.}
\crefname{table}{Table}{Tables}
\crefname{figure}{Fig.}{Figs.}

\begin{document}
\title{\paperTitle}
\author{\authorBlock}
\maketitle

\begin{abstract}
Domain generalization (DG) is proposed to deal with the issue of domain shift, which occurs when statistical differences exist between source and target domains. However, most current methods do not account for a common realistic scenario where the source and target domains have different classes. To overcome this deficiency, open set domain generalization (OSDG) then emerges as a more practical setting to recognize unseen classes in unseen domains. An intuitive approach is to use multiple one-vs-all classifiers to define decision boundaries for each class and reject the outliers as unknown. However, the significant class imbalance between positive and negative samples often causes the boundaries biased towards positive ones, resulting in misclassification for known samples in the unseen target domain. In this paper, we propose a novel meta-learning-based framework called dualistic MEta-learning with joint DomaIn-Class matching (MEDIC), which considers gradient matching towards inter-domain and inter-class splits simultaneously to find a generalizable boundary balanced for all tasks. Experimental results demonstrate that MEDIC not only outperforms previous methods in open set scenarios, but also maintains competitive close set generalization ability at the same time. Our code is available at \href{https://github.com/zzwdx/MEDIC}{https://github.com/zzwdx/MEDIC}.

\end{abstract}

\section{Introduction}
\label{sec:intro}
Deep neural networks have achieved enormous success in a wide range of computer vision tasks, usually assuming that the training and test samples are drawn from the same data distribution and label space.
However, due to the unpredictability of real-world application scenarios, the model faces the risk of performance degradation when the above constraints are not satisfied \cite{li2019research}. 
Domain generalization (DG) \cite{wang2022generalizing} is then motivated as a more realistic setting to deal with data distribution shift, which refers to using multiple source domains via data augmentation \cite{zhou2020domain, li2021simple}, feature alignment \cite{dou2019domain, li2018domain}, meta-learning \cite{li2018learning, li2019episodic} and so on, to obtain a model with the generalization ability that can be directly applied to arbitrary unseen target domains.

\begin{figure}[tp]
    \centering
    \includegraphics[width=1\linewidth]{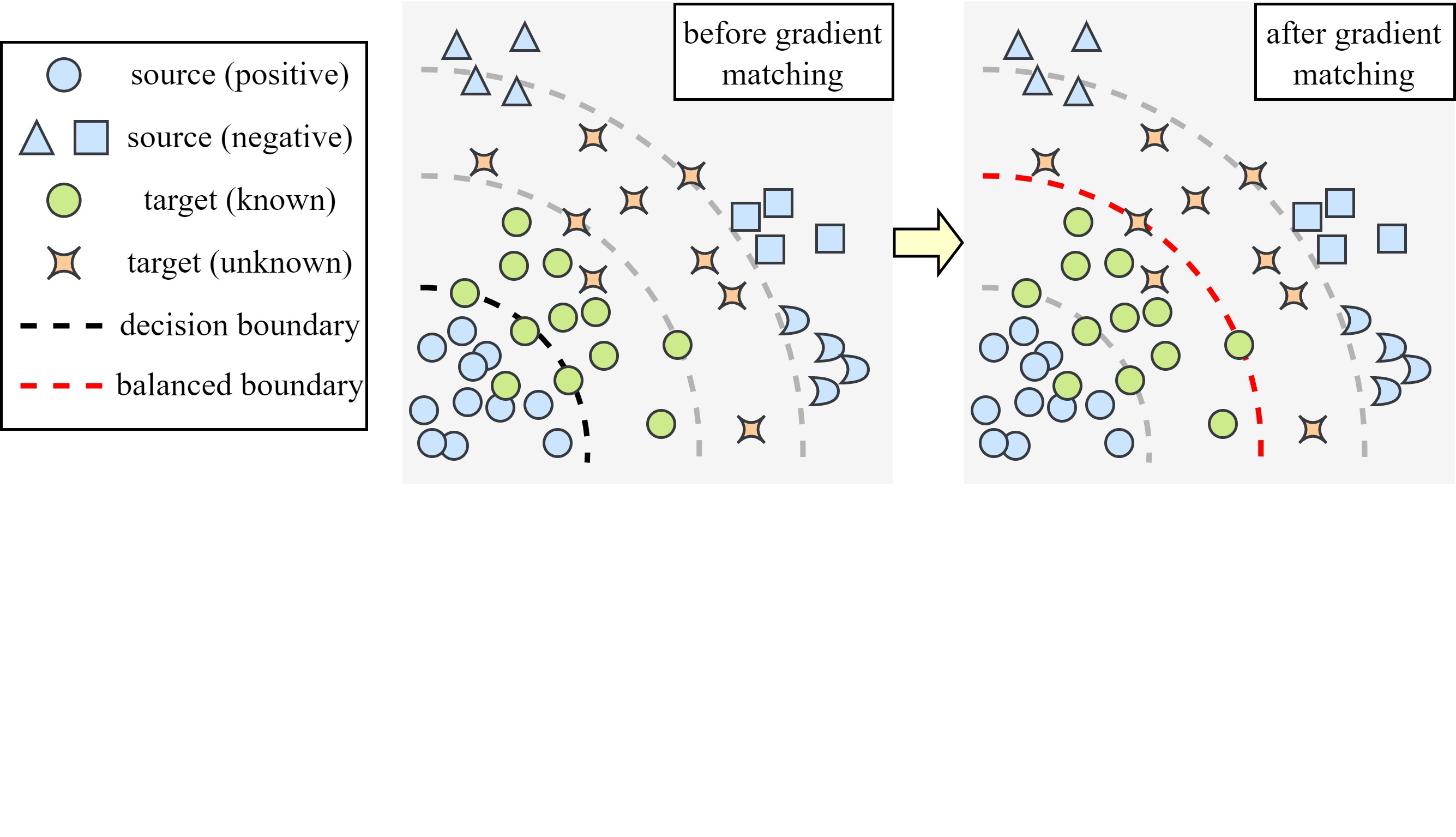}
    \vspace{-1.02in}
    \caption{An example for the variation of decision boundaries of a \emph{one-vs-all} classifier in open set domain generalization.}
    \vspace{-0.1in}
    \label{fig:bias}
\end{figure}

Most current domain generalization researches are based on the assumption of close set recognition, \ie, the classes in the source domains are consistent with that of the target domains. 
However, in practice, the deployed model is often exposed to some new classes that have never been encountered in the training phase \cite{scheirer2012toward}.
For example, in self-driving tasks, thousands of objects are fed into the model that the known label space cannot guarantee complete coverage of them; in medical image processing tasks, some diseases are extremely rare that it is unrealistic to acquire their samples for training.
According to the logic of close set classification, these potential objects are forced to be identified into a known class, which lays hidden dangers for the model's robustness and security.
To address this problem, it is necessary to investigate a more practical setting of open set domain generalization (OSDG), which focuses on recognizing unknown classes without losing the original classification accuracy.

In open set domain generalization \cite{shu2021open, katsumata2021open}, the key is to concurrently deal with the problems of domain shift and category shift. 
We now turn our attention to a simple yet elegant learning paradigm, meta-learning \cite{hospedales2021meta}, which has shown its effectiveness in addressing domain shift.
Previous research on meta-learning-based DG \cite{li2018learning} and OSDG \cite{shu2021open} aims to find an optimal balance between different domains through implicit gradient matching between tasks sampled from them. This domain-wise meta-learning helps prevent the model from becoming overly biased towards individual ones.
The rest issue is how to effectively model the category shift. A natural idea is to use a multi-binary classifier \cite{saito2021ovanet, liu2019separate}, which comprises multiple one-vs-all binary classifiers, to define a decision boundary for each known class. If a given sample is classified as negative by all sub-classifiers, it is considered to have a high probability of belonging to the unknown classes. 
However, we found this issue is non-trivial because the one-vs-all classifier in multi-classification tasks can generate biased predictions.
As shown in \cref{fig:bias}, due to the limited data distribution of positive samples (\ie, from the corresponding class) and the diverse data distribution of negative samples (\ie, from all of the other classes), the classifier may asymmetrically focus on the negative ones which are easier to optimize.
This can cause the decision boundary to cling to the positive samples, leading to inaccurate identification of known class samples in the unseen target domain.
Conversely, if the decision boundary is too far away from the positive samples, the model can be prone to predicting an excessive quantity of target samples as known, hindering the recognition performance of unknown classes.

To establish a well-balanced decision boundary between classes, we attempt to fully exploit the gradient matching property of meta-learning.
In contrast to the domain-wise strategy, we additionally sample tasks by classes to make the decision boundary rationally lying at the middle zone of each class.
As a result, unknown samples are more likely to be distributed near the decision boundary, making them easier to be recognized.
Concretely, we mitigate the limitation in previous works that merely select tasks by domains \cite{shi2021gradient, rame2022fishr} and propose a novel meta-learning strategy called \emph{dualistic MEta-learning with joint DomaIn-Class matching (MEDIC)}.  
For tasks selected from different domains, we further divide them at category level.
By matching the gradients of these divided tasks, we expect the model not only to generalize well across domains, but also to grasp the class-wise relationship more precisely.
Furthermore, with the help of an auxiliary multi-binary classifier \cite{liu2019separate, saito2021ovanet}, the model can automatically learn decision boundaries for each individual class.

\section{Related Work}
\label{sec:related}

\textbf{Domain Generalization (DG)} 
is intended to learn a model from multiple source domains that can migrate to any unseen target domain with no additional retraining process.
Existing domain generalization methods can be roughly divided into three genres:
(i) \textbf{Feature Representation} \cite{kim2021selfreg} aims to extract universal features across source domains, assuming they are also applicable to the target domain, such as domain adversarial learning \cite{li2018domain, ganin2016domain} which trains a domain-irrelevant feature extractor guided by a domain-sensitive discriminator, feature decoupling \cite{chattopadhyay2020learning} that distills domain-invariant features from the original ones.
(ii) \textbf{Data Augmentation.} 
The goal of data augmentation is to further increase the diversity of training data to enhance the generalization ability of the model.
For example, \cite{zhou2020domain, guo2023aloft} focus on mixing or perturbing statistics at image or feature level, while mapping data distribution \cite{zhou2020learning, li2021progressive} and injecting random noise \cite{li2021simple} are also proposed to generate artificial domains.
(iii) \textbf{Learning Strategy.}
Meta-learning \cite{zhang2022mvdg, dou2019domain,  balaji2018metareg} is aimed at leveraging past knowledge to improve the learning of current tasks. 
For example, MLDG \cite{li2018learning} simulates domain shift by synthesizing meta-train and meta-test domains, which exerts a far-reaching influence on the following meta-learning-based researches.
Ensemble learning \cite{zhou2021domain, cha2021swad} attempts to reduce the risk of overfitting by combining models learned from different domains or sampled from a series of episodes.
Gradient-based strategies perform regularization by shielding some gradients at each iteration \cite{huang2020self} or intervening in the direction of gradient update \cite{shi2021gradient, mansilla2021domain} which sometimes can be indirectly achieved through meta-learning.

\begin{table}[t]
\caption{Comparison of target domains under different problem settings. The data distribution and label space of source domains are expressed as $\mathcal{P}$ and $\mathcal{C}$ respectively. The data distribution $\mathcal{Q}$ is unseen and the label space $\mathcal{U}$ satisfies $\mathcal{C}\cap \mathcal{U}=\varnothing$.}
\vspace{-0.05in}
\centering
\resizebox{1.\linewidth}{!}{
\begin{tabular}{lccc}
\toprule
\textbf{\makecell[l]{Problem Setting}} & 
{\makecell[c]{Distribution \\ of Data}} & 
{\makecell[c]{Label \\ Space}} & 
{\makecell[c]{Participation \\ in Training}} \\
\midrule
Domain Adaptation \cite{wang2018deep} & $\mathcal{Q}$ & $\mathcal{C}$ & \checkmark \\
Domain Generalization \cite{wang2022generalizing} & $\mathcal{Q}$ & $\mathcal{C}$ & $\times$ \\
Open Set Recognition \cite{geng2020recent} & $\mathcal{P}$ & $\mathcal{C \cup U}$ & $\times$ \\
Open Set Domain Generalization \cite{shu2021open} & $\mathcal{Q}$ & $\mathcal{C\cup U}$ & $\times$ \\
\bottomrule
\end{tabular}
}

\vspace{-0.1in}
\label{tab: pbs-cmp}
\end{table}

\textbf{Open Set Recognition (OSR)} aims to detect novel classes not existing in the training set.
From the perspective of whether to use extra data, current open set recognition methods can be classified into two categories.
(i) \textbf{Artificial Classes.}
\cite{dhamija2018reducing, hendrycks2018deep} propose to enrich the training set with classes of no interest to facilitate the feature separation of known classes.
However, the effectiveness of these methods heavily relies on the quality of auxiliary samples and it may not be fair to compare them with other methods in a unified framework.
Other methods \cite{ge2017generative, neal2018open} adopt generative adversarial networks (GAN) to automatically generate samples of virtual unknown classes.
(ii) \textbf{Discriminative Models.}
OpenMax \cite{bendale2016towards} abandons the traditional softmax layer and estimates the unknown probability of each sample with extreme value theory (EVT).
Based on the assumption that the reconstructions of known class samples are usually more accurate, \cite{oza2019c2ae, yoshihashi2019classification} utilize reconstruction error to find the threshold for determining unknown classes.
The idea of metric learning \cite{chen2021adversarial, guo2021conditional} that aggregates intra-class and separate inter-class samples has also been applied to obtain models with higher discrimination ability.
However, the above methods tend to treat all out-of-distribution samples as unknown classes, making it challenging to directly apply them to the DG setting.

\textbf{Open Set Domain Generalization (OSDG)}, which is listed with other problem settings in \cref{tab: pbs-cmp}, is still in its early stages.
There are very few related works as far as we know.
\cite{shu2021open} proposes a DAML framework based on domain augmentation and meta-learning. 
Since the model is not explicitly or implicitly informed by the existence of potential unknown classes in both training and inference phases, this work is in a way more focused on close set domain generalization instead.
\cite{katsumata2021open} borrows metric learning to diffuse the feature representations of unknown classes but relies on existing DG baselines to acquire domain-invariant features, so we think it should be more regarded as an OSR method with DG as a special case.
In addition, we find two recent studies \cite{zhu2021crossmatch, yang2022one} in open set single domain generalization (OS-SDG), which differs from OSDG in that only one source domain is available for training.
\cite{zhu2021crossmatch} recommends imitating unknown classes via adversarial learning, 
and \cite{yang2022one} forges unknown class samples by masking the labels of original ones.
To sum up, the above methods usually have a sense of fragmentation when processing domain and category shifts, while our goal is to integrate them seamlessly.

\section{Method}
\label{sec:method}

\subsection{Problem Analysis}
\label{subsec:mpa}

In open set domain generalization, we are provided with source domains $\mathcal{S}=\lbrace \mathcal{D}_1,\mathcal{D}_2,...,\mathcal{D}_S\rbrace$ with a label space $\mathcal{C}$ and unseen target domains $\mathcal{T}=\lbrace \mathcal{D}_{S+1},\mathcal{D}_{S+2},...,\mathcal{D}_{S+T}\rbrace$ with an extended label space $\mathcal{C}\cup\mathcal{U}$ that ensure $\mathcal{C}\cap \mathcal{U}=\varnothing$. The $s$-th domain consisting of $N_s$ samples is represented as $\mathcal{D}_s=\lbrace(x^s_i,y^s_i)\rbrace_{i=1}^{N_s}$, where $x^s_i$ denotes the $i$-th sample from the sample space $\mathcal{X}$ and $y^s_i$ can take values in $\mathcal{C}$ or $\mathcal{C}\cup\mathcal{U}$, signifying the corresponding label in the source or target domains.
Our goal is to fully exploit these source domains $\mathcal{S}$ to obtain a model that can directly generalize to any unseen target domain with unknown classes. 
The meta-learning \cite{li2018learning} training scheme requires to split the source domains $\mathcal{S}$ into meta-train set $\mathcal{S}_{\mathcal{F}}$ and meta-test set $\mathcal{S}_{\mathcal{G}}$, where $\mathcal{S}_{\mathcal{F}} \cup \mathcal{S}_{\mathcal{G}}=\mathcal{S}$ and $\mathcal{S}_{\mathcal{F}} \cap \mathcal{S}_{\mathcal{G}}=\varnothing$. We define the loss of the model trained on the data sampled from these two sets $\mathcal{S}_{\mathcal{F}}$ and $\mathcal{S}_{\mathcal{G}}$ as $\mathcal{F}(\Theta)$ and $\mathcal{G}(\Theta)$, respectively, where $\Theta$ is the parameter of a training model. 
First, we update the model's parameter to $\hat{\Theta}$ with the loss of $\mathcal{F}(\Theta)$. 
Then, we utilize the loss of $\mathcal{G}(\hat{\Theta})$ to update its original parameter $\Theta$.

\begin{figure}[t]

\vspace{-0.15in}
\begin{algorithm}[H]
\label{alg:alg}
\caption{Training process of MEDIC}  
\begin{algorithmic}[1]    
\renewcommand{\algorithmicrequire}{\textbf{Input:}}
\REQUIRE Domains $\mathcal{S}$; classes $\mathcal{C}$; model parametrized by $\Theta$; loss function $\mathcal{L}$; learning rates $\alpha, \beta$ and $\eta$
\WHILE{$\Theta$ not converged}
\STATE \textbf{Random split} $\mathcal{S}_1, \mathcal{S}_2 \leftarrow \mathcal{S}$; $\mathcal{C}_1, \mathcal{C}_2 \leftarrow \mathcal{C}$
\STATE \textbf{Sample} ${\mathcal{B}}_{\mathcal{F}_1}$ from $(\mathcal{S}_1, \mathcal{C}_1)$; ${\mathcal{B}}_{\mathcal{F}_2}$ from $(\mathcal{S}_1, \mathcal{C}_2)$; ${\mathcal{B}}_{\mathcal{G}_1}$ from $(\mathcal{S}_2, \mathcal{C}_1)$; ${\mathcal{B}}_{\mathcal{G}_2}$ from $(\mathcal{S}_2, \mathcal{C}_2)$
\STATE \textbf{Meta-train:} $\mathcal{L}_1 \leftarrow \mathcal{L}({\mathcal{B}}_{\mathcal{F}_1}; \Theta) + \mathcal{L}({\mathcal{B}}_{\mathcal{G}_2}; \Theta)$ \\
$\hat{\Theta} \leftarrow \Theta - \alpha \cdot \nabla_\Theta \mathcal{L}_1$
\STATE \textbf{Meta-test:} $\mathcal{L}_2 \leftarrow \mathcal{L}({\mathcal{B}}_{\mathcal{F}_2}; \hat{\Theta}) + \mathcal{L}({\mathcal{B}}_{\mathcal{G}_1}; \hat{\Theta})$ \\
$\Theta \leftarrow \Theta - \eta \cdot (\nabla_\Theta \mathcal{L}_1 + \beta \cdot \nabla_{\hat{\Theta}} \mathcal{L}_2)$

\ENDWHILE

\end{algorithmic}

\end{algorithm}
\vspace{-0.22in}
\end{figure}


\begin{figure*}[tp]  
    \centering
    \vspace{-0.05in}
    \includegraphics[width=1\linewidth]{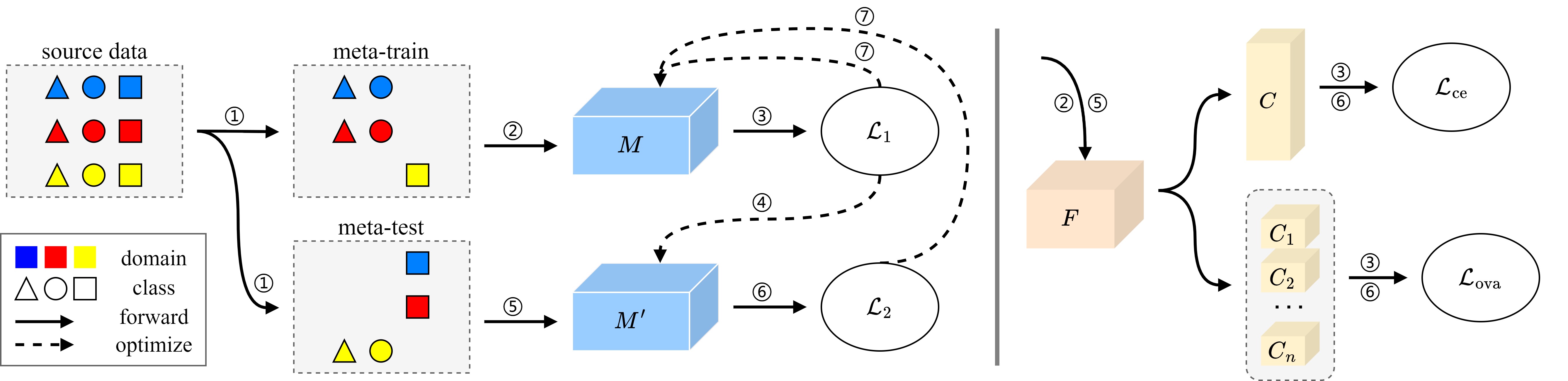}
    \caption{
    Overview of MEDIC during one training iteration. 
    $M$ is the overall model, and the right figure represents its internal structure.
    The numbers denote the sequence of data flow (solid arrows) and model updates (dashed arrows) respectively. 
    } 
    \vspace{-0.1in}
    \label{fig:model}
\end{figure*}

\subsection{The MEDIC Framework}
\textbf{Dualistic Meta-Learning.}
Given two data splits $\mathcal{S}_{\mathcal{F}}$ and $\mathcal{S}_{\mathcal{G}}$ and their corresponding loss function $\mathcal{F}$ and $\mathcal{G}$, our objective is to reach a consensus on their gradients ${\mathcal{F}'(\Theta)}$ and ${\mathcal{G}'(\Theta)}$ to ensure unbiased optimization direction for the prediction of $\mathcal{S}_{\mathcal{F}}$ and $\mathcal{S}_{\mathcal{G}}$. 
The rationale behind it is that if the angle between the directions of ${\mathcal{F}'(\Theta)}$ and ${\mathcal{G}'(\Theta)}$ is small which means optimizing one task does not hurt the other task, then updating with their combined gradient (\ie, summed gradients in practice) can result in a better performance. Instead, if the angle between them is large which means these two tasks conflict with each other, the optimization procedure will be damaged and lead to an inferior optimization process.
The key idea of gradient matching is to find a position in the weight space where the angle between the gradient of $\mathcal{S}_{\mathcal{F}}$ and $\mathcal{S}_{\mathcal{G}}$ is small, which is equivalent to maximize the dot product of ${\mathcal{F}'(\Theta)}$ and ${\mathcal{G}'(\Theta)}$~\cite{li2018learning}. After moving the model towards this area in the last training step, $\mathcal{S}_{\mathcal{F}}$ and $\mathcal{S}_{\mathcal{G}}$ will reach an agreement on the direction of gradient update.

Current gradient-based domain generalization methods often split $\mathcal{S}_{\mathcal{F}}$ and $\mathcal{S}_{\mathcal{G}}$ into different domains to find an optimization direction that is invariant only among domains \cite{li2018learning}. However, these methods ignore the crucial relationship between classes which is necessary for open set recognition. 
Instead of simply adding an extra iteration to avoid biased prediction between classes, we take a step further and propose a novel meta-learning strategy named \emph{dualistic MEta-learning with joint DomaIn-Class matching (MEDIC)}, which achieves gradient matching towards inter-domain and inter-class splits simultaneously to learn a more generalizable decision boundary balanced for all tasks.

As illustrated in \cref{fig:model}, for $\mathcal{S}_{\mathcal{F}}$ and $\mathcal{S}_{\mathcal{G}}$ derived from disparate source domains, we further divide them into $\mathcal{S}_{\mathcal{F}_1}, \mathcal{S}_{\mathcal{F}_2}$ and $\mathcal{S}_{\mathcal{G}_1}, \mathcal{S}_{\mathcal{G}_2}$ by classes and define their corresponding loss functions as $\mathcal{F}_1, \mathcal{F}_2$ and $\mathcal{G}_1, \mathcal{G}_2$. The label spaces between $\mathcal{S}_{\mathcal{F}_1}$ and $\mathcal{S}_{\mathcal{F}_2}$ as well as between $\mathcal{S}_{\mathcal{G}_1}$ and $\mathcal{S}_{\mathcal{G}_2}$ are both disjoint.
Besides, we require that $\mathcal{S}_{\mathcal{F}_1}$ and $\mathcal{S}_{\mathcal{G}_1}$ share the same labels, and likewise for $\mathcal{S}_{\mathcal{F}_2}$ and $\mathcal{S}_{\mathcal{G}_2}$. 
Then to implement the gradient matching between domains and classes simultaneously, we employ $(\mathcal{S}_{\mathcal{F}_1}, \mathcal{S}_{\mathcal{G}_2})$ as meta-train set and $(\mathcal{S}_{\mathcal{F}_2}, \mathcal{S}_{\mathcal{G}_1})$ as meta-test set respectively.  
The final meta-objective function of MEDIC is defined as follows:
\begin{equation}
\label{eq:objective}
\mathop{\rm argmin}_{\Theta}\, [{\mathcal{F}_1(\Theta)} + {\mathcal{G}_2(\Theta)}] + \beta
[{\mathcal{F}_2(\hat{\Theta})} + {\mathcal{G}_1(\hat{\Theta})}],
\end{equation}
where $\beta$ measures the weight between the two meta sets and $\hat{\Theta}$ is the optimized model parameters on the meta-train set with learning rate $\alpha$: 
\begin{equation}
\label{eq:thetahat}
{\hat{\Theta}} = \Theta - \alpha({\mathcal{F}_1'(\Theta)} + {\mathcal{G}_2'(\Theta)}).
\end{equation}
To verify MEDIC can exactly achieve gradient matching between domains and classes at the same time, similar to the analysis in \cite{li2018learning}, we perform the first order Taylor expansion for the second term in \cref{eq:objective}:
\begin{align}
    &{\mathcal{F}_2(\hat{\Theta})} = {\mathcal{F}_2(\Theta)} - \alpha (
{\mathcal{F}_1'(\Theta)} + {\mathcal{G}_2'(\Theta)}) \cdot {\mathcal{F}_2'(\Theta)}, \\
    &{\mathcal{G}_1(\hat{\Theta})} = {\mathcal{G}_1(\Theta)} - \alpha (
{\mathcal{F}_1'(\Theta)} + {\mathcal{G}_2'(\Theta)}) \cdot {\mathcal{G}_1'(\Theta)},
\end{align}
and the objective function becomes:
\begin{equation}
\label{eq:taylorobjective}
\begin{split}
& \mathop{\rm argmin}_{\Theta}\, [{\mathcal{F}_1(\Theta)} + {\mathcal{G}_2(\Theta)} + \beta
({\mathcal{F}_2(\Theta)} +  {\mathcal{G}_1(\Theta)})] \\ &- \beta \alpha [({\mathcal{F}_1'(\Theta)} + {\mathcal{G}_2'(\Theta)})\cdot(
{\mathcal{F}_2'(\Theta)} +
{\mathcal{G}_1'(\Theta)})].
\end{split}
\end{equation}
The first term of \cref{eq:taylorobjective} means optimizing the model with their original losses. The second term is the sum of the product of gradients. By expanding this term, we can obtain the following regularization term:
\begin{equation}
\label{eq:regularization}
\mathcal{L}_{\rm reg} = -(\mathcal{F}_1' \cdot \mathcal{F}_2' + 
\mathcal{F}_1' \cdot
\mathcal{G}_1' +
\mathcal{G}_2' \cdot
\mathcal{F}_2' +
\mathcal{G}_2' \cdot
\mathcal{G}_1').
\end{equation}
As previously mentioned, maximizing the dot product of gradients can regularize the training procedure that matches the updating directions of different tasks.
Since for $\mathcal{S}_{\mathcal{F}_1}$, the counterpart $\mathcal{S}_{\mathcal{G}_1}$ contains the same classes in different domains, while $\mathcal{S}_{\mathcal{F}_2}$ contains different classes in the same domain, the two terms in any one of the above gradient products are either from different domains or different classes, allowing for domain-wise and class-wise matching simultaneously.
Compared with conventional methods which only consider inter-domain relationships, the dot product between $\mathcal{S}_{\mathcal{F}_1}$ and $\mathcal{S}_{\mathcal{F}_2}$ fills the gap of class-wise gradient matching inside each domain to learn a more reasonable decision boundary, achieving a more fine-grained model optimization.

\textbf{Open Set Loss Function.}
For open set recognition, we adopt another multi-binary classifier \cite{saito2021ovanet} to serve as a supplement to conventional close set classifier.
As illustrated in \cref{fig:model}, the proposed classifier consists of $|\mathcal{C}|$ one-vs-all classifiers, where each classifier is trained to detect whether a sample belongs to the corresponding class.
Let $p(\hat{y}^k|x)$ denote the output probability that an instance $x$ is an inlier of the $k$-th sub-classifier.
For a given sample $(x, y)$, its loss on the multi-binary classifier can be presented as:
\begin{equation}
\label{eq:ovaloss}
\mathcal{L}_{\rm ova}(x, y) = -{\rm log}(p(\hat{y}^y|x)) - \mathop{\rm min}_{j \neq y}\,{\rm log}(1-p(\hat{y}^j|x)).
\end{equation}
The second term denotes that it updates only the most challenging binary classifier when used as a negative sample.
We simply adopt this loss and train a close set classifier using cross-entropy loss which is denoted as $\mathcal{L}_{\rm ce}$.
Then the overall open set loss function can be defined as follows:
\begin{equation}
\label{eq:overalloss}
\mathcal{L}_{\rm all} = \mathcal{L}_{\rm ce} + \mathcal{L}_{\rm ova},
\end{equation}
which is equivalent to the meta-objective of $\mathcal{F}_1, \mathcal{F}_2, \mathcal{G}_1, \mathcal{G}_2$.
The optimization of the multi-binary classifier is facilitated by MEDIC. Although the proposed classifier can learn a boundary between inliers and outliers for each known class, negative training samples for a certain sub-classifier differ a lot among $|\mathcal{C}|-1$ classes, which may lead to confusion in the direction of gradient descent, causing damage to the gradient of positive ones. 
The application of inter-class gradient alignment can stabilize both of them, thus seeking a balance between close set generalization and open set recognition. 

\subsection{Inference}
In the test phase, each target sample is first predicted by the close set classifier to obtain a probability distribution $p(\hat{y}|x)$ among known classes.
The model either (i) selects the value of its maximum likelihood with the original classifier:
\begin{equation}
\label{eq:confcls}
{\rm conf}_{\rm cls}(x) = {\rm max}_{i=1}^{|\mathcal{C}|}(p(\hat{y}|x)_i),
\end{equation}
or (ii) refers to the corresponding one-vs-all classifier and choose the value of its positive output channel as the confidence score \cite{saito2021ovanet}: 
\begin{equation}
\label{eq:confblcs}
{\rm conf}_{\rm bcls}(x) = p(\hat{y}^{{\rm argmax}_{i=1}^{|\mathcal{C}|}(p(\hat{y}|x)_i)}|x).
\end{equation}
If the score is larger than a preset threshold $\mu$, then detect the sample as known and generate a particular class label, otherwise judge it as unknown.
Experimental results are reported on these two inference modes in \cref{sec:experiment}.
We use cls and bcls to express discriminating unknown classes with close set classifier and multi-binary classifier respectively.

\begin{table*}[t]
\vspace{-0.1in}
\caption{Results (\%) of PACS on ResNet18.
The ratio of known to unknown classes is 6:1. (Best in \textbf{bold})}
\vspace{-0.05in}
\centering
\resizebox{1.\linewidth}{!}{
\begin{tabular}{l|ccc|ccc|ccc|ccc|ccc}
\toprule
& \multicolumn{3}{c|}{\textbf{Photo}} & \multicolumn{3}{c|}{\textbf{Art}} & \multicolumn{3}{c|}{\textbf{Cartoon}} & \multicolumn{3}{c|}{\textbf{Sketch}} & \multicolumn{3}{c}{\textbf{Avg}} \\
\textbf{Method} & Acc & H-score & OSCR & Acc & H-score & OSCR & Acc & H-score & OSCR & Acc & H-score & OSCR & Acc & H-score & OSCR \\

\midrule
OpenMax\footnotemark[2] \cite{bendale2016towards} & 95.56 & 92.48 & - & 83.68 & 69.61 & - & 78.61 & 64.36 & - & 70.89 & 50.67 & - & 82.19 & 69.28 & - \\

ERM \cite{naumovich1998statistical} & \textbf{96.04} & 93.40 & \textbf{95.11} & 84.18 & 70.54 & 71.89 &  77.63 & 62.80 & 62.57 & 70.44 & 55.81 & 51.75 & 82.07 & 70.64 & 70.33 \\

ARPL\footnotemark[2]  \cite{chen2021adversarial} & 94.83 & \textbf{95.06} & 94.63 & 83.93 & 67.88 & 68.82 & 78.56 & 62.98 & 65.30 & 74.34 & 61.20 & 59.80 & 82.91 & 71.78 & 72.14 \\

MMLD \cite{matsuura2020domain} & 94.83 & 88.80 & 92.94 & 84.43 & 64.83 & 69.43 & 77.11 & 64.21 & 65.36 & 75.14 & 67.70 & 64.69 & 82.88 & 71.38 & 73.11 \\

RSC \cite{huang2020self} & 94.43 & 88.37 & 91.38 & 83.36 & 70.27 & 73.55 & 78.09 & 65.13 & 66.15 & 77.16 & 52.98 & 62.31 & 83.26 & 69.19 & 73.35 \\


DAML\footnotemark[2]  \cite{shu2021open} & 91.44 & 80.87 & 82.83 & 83.11 & 72.05 & 71.75 & 79.11 & \textbf{66.26} & 66.46 & \textbf{82.97} & 72.63 & 73.71 & 84.16 & 72.95 & 73.69 \\

MixStyle \cite{zhou2020domain} & 95.23 & 82.02 & 88.99 & 86.18 & 70.62 & 72.57 & 78.92 & 63.23 & 63.81 & 80.34 & 71.90 & 72.07 & 85.17 & 71.94 & 74.36 \\

SelfReg \cite{kim2021selfreg} & 95.72 & 89.34 & 92.26 & 86.24 & 72.45 & 73.77 & 80.77 & 65.75 & 66.38 & 78.30 & 67.06 & 65.69 & 85.26 & 73.65 & 74.53 \\

MLDG \cite{li2018learning} & 94.99 & 91.48 & 93.70 & 84.12 & 69.52 & 72.15 & 78.45 & 61.59 & 64.32 & 79.99 & 69.67 & 68.60 & 84.39 & 73.06 & 74.69\\

MVDG \cite{zhang2022mvdg} & 94.43 & 74.07 & 88.07 & \textbf{87.62} & 71.98 & 75.05 & 81.18 & 63.95 & 66.34 & 82.41 & 73.55 & 73.83 & \textbf{86.41} & 70.89 & 75.82 \\

\midrule
MEDIC-cls & 94.83 & 83.68 & 90.30 & 86.20 & 69.35 & 74.16 & 81.94 & 63.26 & 67.43 & 81.84 & 69.60 & 70.85 & 86.20 & 71.47 & 75.69 \\

MEDIC-bcls & 94.83 & 89.49 & 92.40 & 86.20 & \textbf{73.82} & \textbf{75.58} & \textbf{81.94} & \textbf{66.26} & \textbf{69.04} & 81.84 & \textbf{74.37} & \textbf{74.52} & 86.20 & \textbf{75.98} & \textbf{77.89} \\

\bottomrule
\end{tabular}
}
\vspace{-0.05in}
\label{tab: pacs6-res18}
\end{table*}
\begin{table*}[t]
\caption{Results (\%) of PACS on ResNet50.
The ratio of known to unknown classes is 6:1. (Best in \textbf{bold})}
\vspace{-0.05in}
\centering
\resizebox{1.\linewidth}{!}{
\begin{tabular}{l|ccc|ccc|ccc|ccc|ccc}
\toprule
& \multicolumn{3}{c|}{\textbf{Photo}} & \multicolumn{3}{c|}{\textbf{Art}} & \multicolumn{3}{c|}{\textbf{Cartoon}} & \multicolumn{3}{c|}{\textbf{Sketch}} & \multicolumn{3}{c}{\textbf{Avg}} \\
\textbf{Method} & Acc & H-score & OSCR & Acc & H-score & OSCR & Acc & H-score & OSCR & Acc & H-score & OSCR & Acc & H-score & OSCR \\

\midrule
OpenMax\footnotemark[2]  \cite{bendale2016towards} & \textbf{97.58} & 93.09 & - & 88.37 & 73.91 & - & 84.38 & 68.23 & - & 80.07 & 68.06 & - & 87.60 & 75.82 & - \\

ARPL\footnotemark[2]  \cite{chen2021adversarial} & 97.09 & \textbf{96.81} & \textbf{96.86} & 88.24 & 77.48 & 80.32 & 82.68 & 67.19 & 68.31 & 78.08 & 70.04 & 69.47 & 86.52 & 77.88 & 78.74\\

MIRO \cite{cha2022domain} & 94.85 & 92.32 & 93.27 & 88.51 & 65.02 & 79.01 & 82.98 & 63.05 & 73.72 & 82.22 & 69.47 & 70.61 & 87.14 & 72.47 & 79.15 \\

MLDG \cite{li2018learning} & 96.77 & 95.85 & 96.33 & 87.99 & 77.16 & 79.93 & 83.45 & 68.74 & 71.32 & 82.25 & 73.16 & 72.27 & 87.61 & 78.73 & 79.96\\

ERM \cite{naumovich1998statistical} & 97.09 & 96.58 & 96.68 & 89.99 & 76.05 & 82.44 & 85.10 & 65.79 & 70.59 & 80.31 & 70.29 & 70.16 & 88.12 & 77.18 & 79.97 \\

CIRL \cite{lv2022causality} & 96.53 & 87.75 & 95.40 & 92.06 & 70.75 & 77.44 & 85.71 & 68.82 & 73.71 & 84.35 & 66.73 & 77.24 & 89.66 & 73.51 & 80.95 \\

MixStyle \cite{zhou2020domain} & 96.53 & 93.57 & 95.30 & 90.87 & 79.15 & 83.27 & \textbf{86.80} & 68.08 & 74.68 & \textbf{84.88} & 71.57 & 73.41 & 89.77 & 78.09 & 81.66\\

CrossMatch\footnotemark[2]  \cite{zhu2021crossmatch} & 96.53 & 96.34 & 96.12 & 91.37 & 75.67 & 82.32 & 83.92 & 67.02 & 74.55 & 81.61 & 72.03 & 73.99  & 88.37 & 77.76 & 81.75 \\

SWAD \cite{cha2021swad} & 96.37 & 84.56 & 93.24 & \textbf{93.75} & 68.41 & 85.00 & 85.57 & 58.57 & 75.90 & 81.90 & 74.66 & 74.65 & 89.40 & 71.55 & 82.20 \\

MVDG \cite{zhang2022mvdg} & 97.17 & 95.02 & 96.63 & 92.50 & 79.47 & 85.02 & 86.02 & 71.05 & 76.03 & 83.44 & 75.24 & 75.18 & 89.78 & 80.20 & 83.21 \\ 

\midrule
MEDIC-cls & 96.37 & 93.80 & 95.37 & 91.62 & 80.80 & 84.67 & 86.65 & 75.85 & 77.48 & 84.61 & 75.80 & 76.79 & 89.81 & 81.56 & 83.58 \\

MEDIC-bcls & 96.37 & 94.75 & 95.79 & 91.62 & \textbf{81.61} & \textbf{85.81} & 86.65 & \textbf{77.39} & \textbf{78.30} & 84.61 & \textbf{78.35} & \textbf{79.50} & \textbf{89.81} & \textbf{83.03} & \textbf{84.85} \\

\bottomrule
\end{tabular}
}
\vspace{-0.05in}
\label{tab: pacs6-res50}
\end{table*}
\begin{table*}[t]
\caption{Results (\%) of Digits-DG on ConvNet.
The ratio of known to unknown classes is 6:4. (Best in \textbf{bold})} 
\vspace{-0.05in}
\centering
\resizebox{1.\linewidth}{!}{
\begin{tabular}{l|ccc|ccc|ccc|ccc|ccc}
\toprule
& \multicolumn{3}{c|}{\textbf{MNIST}} & \multicolumn{3}{c|}{\textbf{MNIST-M}} & \multicolumn{3}{c|}{\textbf{SVHN}} & \multicolumn{3}{c|}{\textbf{SYN}} & \multicolumn{3}{c}{\textbf{Avg}} \\
\textbf{Method} & Acc & H-score & OSCR & Acc & H-score & OSCR & Acc & H-score & OSCR & Acc & H-score & OSCR & Acc & H-score & OSCR \\

\midrule
OpenMax\footnotemark[2]  \cite{bendale2016towards} & 97.33 & 52.03 & - & 71.03 & 57.26 & - & 72.00 & 49.46 & - & 84.83 & 54.78 & - & 81.30 & 53.38 & - \\ 

MixStyle \cite{zhou2020domain} & 97.86 & 73.25 & 89.36 & 74.50 & 59.30 & 56.95 & 69.28 & 53.24 & 48.43 & 85.06 & 60.22 & 65.44 & 81.68 & 61.50 & 65.05\\

ERM \cite{naumovich1998statistical} & 97.47 & 80.90 & 92.60 & 71.03 & 53.92 & 54.04 & 71.08 & 54.37 & 49.86 & 85.67 & 51.57 & 67.63 & 81.31 & 60.19 & 66.03\\

ARPL\footnotemark[2]  \cite{chen2021adversarial} & 97.75 & \textbf{85.74} & 91.86 & 69.78 & 58.08 & 54.21 & 71.78 & 56.98 & 53.63 & 85.31 & \textbf{64.04} & 65.89 & 81.16 & 66.21 & 66.40\\

MLDG \cite{li2018learning} & 97.83 & 80.36 & 94.28 & 71.11 & 46.84 & 55.17 & 73.64 & 53.54 & 53.64 & 86.08 & 63.56 & 70.34 & 82.16 & 61.08 & 68.36\\

SWAD \cite{cha2021swad} & 97.71 & 84.44 & 92.65 & \textbf{73.09} & 53.35 & 55.94 & \textbf{76.08} & \textbf{59.18} & 56.25 & 87.95 & 51.27 & 69.03 & \textbf{83.71} & 62.06 & 68.47\\

\midrule

MEDIC-cls & 97.89 & 67.37 & \textbf{96.17} & 71.14 & 48.44 & 55.37 & 76.00 & 51.20 & 55.58 & 88.11 & 64.90 & \textbf{73.62} & 83.28 & 57.98 & 70.19\\

MEDIC-bcls & \textbf{97.89} & 83.20 & 95.81 & 71.14 & \textbf{60.98} & \textbf{58.28} & 76.00 & 58.77 & \textbf{57.60} & \textbf{88.11} & 62.24 & 72.91 & 83.28 & \textbf{66.30} & \textbf{71.15}\\

\bottomrule
\end{tabular}
}
\vspace{-0.1in}                                          
\label{tab: digits6-conv}
\end{table*}

\footnotetext[2]{Open-set-based methods.}

\section{Experiment}
\label{sec:experiment}

\subsection{Setup}
\label{subsec: setup}
We experiment on three standard DG datasets whose details are described as follows:
(i) \textbf{PACS} \cite{li2017deeper} contains four domains (\emph{photo}, \emph{art-painting}, \emph{cartoon}, \emph{sketch}) with a total number of 9,991 pictures. 
All of the domains share the same label space of 7 classes but differ in image style. 
Making use of the official split adopted from \cite{li2017deeper} for training and validation, we evaluate the generalization ability of our model in both close set and open set scenarios.
(ii) \textbf{Office-Home} \cite{venkateswara2017deep} comprises around 15,500 images of 65 classes from four domains (\emph{art}, \emph{clipart}, \emph{product}, \emph{real-world}).
The richness of classes allows us to select the ratio of known to unknown in a relatively flexible way.
With the train-validation split provided by \cite{shu2021open}, we conduct experiments in various known-unknown divisions on this dataset.
(iii) \textbf{Digits-DG} \cite{zhou2020deep} is an aggregation of four classic benchmarks including \emph{mnist}, \emph{mnist-m}, \emph{svhn} and \emph{syn}. 
Digit datasets are popular in OSR since the close set accuracy of the model tends to saturate, making it purer to verify the capability of unknown class recognition.
We utilize the original train-validation split in \cite{zhou2020deep} for our open set domain generalization tasks.

The leave-one-domain-out evaluation protocol is carried out on all benchmarks, \ie, picking one target domain for testing and using the remaining ones for training and validation. 
The classes are then rearranged in alphabetical order, with the former part designated as known classes and the latter part as unknown classes.
The split rate of known and unknown classes on each dataset are explained in the caption of the corresponding table.
For PACS and Office-Home, we employ ResNet18 and ResNet50 
\cite{he2016deep} pretrained on ImageNet \cite{deng2009imagenet} as our backbone networks. 
For Digits-DG which is relatively simple to handle, we borrow a lightweight convolutional architecture called ConvNet from \cite{zhou2020deep}.
The validation set does not contain any unseen classes, and we use close set accuracy for model selection.

\subsection{Evaluation Metrics}
Considering the trade-off between the accuracy of known and unknown classes, we choose three evaluation metrics to verify the performance of our model:
(i) \textbf{Acc} is used to represent the typical close set accuracy.
(ii) \textbf{H-score} \cite{fu2020learning} is the harmonic mean of known class accuracy ${\rm acc}_{\rm k}$ and unknown class accuracy ${\rm acc}_{\rm u}$,
which is widely used in current OSDA and OSDG studies. 
When ${\rm acc}_{\rm k}+{\rm acc}_{\rm u}$ is constant, the smaller the difference between ${\rm acc}_{\rm k}$ and ${\rm acc}_{\rm u}$ is, the larger the H-score will be.
Compared with the weighted average, H-score puts more emphasis on the balance between close set classification and open set recognition.
However, due to the uncertainty of domain shift, the manually designed threshold to reject unknown classes is not applicable for a random target domain with arbitrary image styles.
As shown in \cref{fig:hs-range}, there is a gap between the optimal threshold of the source and target domains.
To mitigate the unreliability of artificial selection, we propose to import a threshold-independent metric (iii) open set classification rate (\textbf{OSCR}) \cite{dhamija2018reducing} which plots the true positive rate against the false positive rate through an ever-moving threshold.
Different from area under the receiver operating characteristic (AUROC) \cite{neal2018open} that ignores
the accuracy of known classes, in OSCR only the correctly labeled samples are considered as true positive ones.

\subsection{Results}
We carry out experiments premised on the existence of unknown target classes.
The results on PACS and Digits-DG are presented in \cref{tab: pacs6-res18}, \cref{tab: pacs6-res50}, and \cref{tab: digits6-conv}.
We first compare our method with DG and OSDG methods such as MLDG \cite{li2018learning}, MVDG \cite{zhang2022mvdg} and DAML \cite{shu2021open}.
Although the accuracy of the known classes is relatively similar, the open set performance of MEDIC with multi-binary classifier is significantly superior to other methods. 
For example, in \cref{tab: pacs6-res18}, MEDIC exceeds the previous OSDG method DAML \cite{shu2021open} by $4.20\%$ on the average of OSCR with even fewer model parameters and a higher close set accuracy, while well ahead of the second best method MVDG \cite{zhang2022mvdg} by $2.07\%$, demonstrating that our method is capable of producing more generalizable and discriminative representations which is beneficial for both DG and OSR tasks.

We also compare with several open set recognition methods such as OpenMax \cite{bendale2016towards} and ARPL \cite{chen2021adversarial}.
Note that we do not calculate the OSCR of OpenMax \cite{bendale2016towards}, because it already possesses a threshold-independent property, \ie, the classifier is designed with $|\mathcal{C}| + 1$ output channels, one of which is corresponding to the probability of the unknown classes.
However, the H-score of OpenMax is still below average, further proving that the way of hard inference derived from source domains is not suitable for the unseen target domains.
It can be observed that ARPL \cite{chen2021adversarial}, which is one of the state-of-the-art approaches for open set recognition, fails to perform well than the methods that are tailored for standard domain generalization.
This may indicate that deep learning models possess a natural inclination to identify unknown classes to some extent and the issue of close set classification under distribution shift remains a top priority in OSDG.

\begin{table}[t]
\caption{Ablation studies (\%) of classifiers and parameter sharing on PACS using ResNet50.
\emph{Dw} and \emph{Cw} refer to domain-wise and class-wise gradient matching that can be achieved by the corresponding method. 
The first three rows of each method denote (i) training and inference with close set classifier only, (ii) training with two classifiers and inference with close set classifier, (iii) training with two classifiers and inference with multi-binary classifier.
The option \emph{share} means sharing parameters between the two classifiers.}
\vspace{-0.05in}
\centering
\resizebox{1.\linewidth}{!}{
\begin{tabular}{lcccccccc}
\toprule
{\textbf{Method}} & Dw & Cw & Option & {\textbf{P}} & {\textbf{A}} & {\textbf{C}} & {\textbf{S}} & {\textbf{Avg}} \\
\midrule
\multicolumn{9}{c}{H-Score} \\
\midrule
& & & - & 96.6 & 76.1 & 65.8 & 70.3 & 77.2 \\
ERM \cite{naumovich1998statistical} & - & - & cls & 97.1 & 77.3 & 64.5 & 69.9 & 77.2\\
& & & bcls & \textbf{97.3} & 79.5 & 71.1 & 71.7 & 79.9\\

\midrule

& & & - & 93.6 & 79.2 & 68.1 & 71.6 & 78.1\\
MixStyle \cite{zhou2020domain} & - & - & cls & 90.9 & 74.8 & 67.6 & 74.8 & 77.0\\
& & & bcls & 91.2 & 78.8 & 71.5 & 77.6 & 79.8\\

\midrule

& & & - & 93.2 & 79.2 & 66.5 & 71.5 & 77.6 \\
- & - & \checkmark & cls & 95.0 & 79.3 & 69.0 & 76.3 & 79.9 \\
& & & bcls & 95.4 & 79.1 & 69.9 & 76.6 & 80.3\\

\midrule

& & & - & 95.8 & 77.2 & 68.7 & 73.2 & 78.7\\
MLDG \cite{li2018learning} & \checkmark & - & cls & 95.8 & 77.9 & 67.2 & 69.4 & 77.6\\
& & & bcls & 95.6 & 80.1 & 72.2 & 71.6 & 79.9\\

\midrule

\multirow{4}{*}{MEDIC} & \multirow{4}{*}{\checkmark} & \multirow{4}{*}{\checkmark} & - & 94.1 & 80.5 & 71.3 & 74.5 & 80.1\\
& & & cls & 93.8 & 80.8 & 75.9 & 75.8 & 81.6\\
& & & bcls & 94.8 & \textbf{81.6} & \textbf{77.4} & \textbf{78.3} & \textbf{83.0}\\
& & & share & 95.6 & 81.0 & 76.0 & 77.3 & 82.5\\
\midrule
\multicolumn{9}{c}{OSCR} \\
\midrule

& & & - & 96.7 & 82.4 & 70.6 & 70.2 & 80.0\\
ERM \cite{naumovich1998statistical} & - & - & cls & \textbf{97.3} & 83.9 & 70.7 & 70.9 & 80.7\\
& & & bcls & 97.1 & 83.8 & 71.1 & 72.0 & 81.0\\

\midrule

& & & - & 95.3 & 83.3 & 74.7 & 73.4 & 81.7\\
MixStyle \cite{zhou2020domain} & - & - & cls & 94.5 & 83.3 & 74.9 & 76.6 & 82.3\\
& & & bcls & 94.4 & 83.3 & 74.1 & 77.4 & 82.3
\\
\midrule

& & & - & 95.2 & 81.8 & 72.6 & 71.5 & 80.3 \\
- & - & \checkmark & cls & 95.7 & 83.9 & 73.7 & 75.8 & 82.3 \\
& & & bcls & 95.6 & 84.1 & 75.0 & 77.0 & 82.9 \\

\midrule

& & & - & 96.3 & 79.9 & 71.3 & 72.3 & 80.0\\
MLDG\cite{li2018learning} & \checkmark & - & cls & 96.7 & 83.1 & 74.6 & 73.5 & 82.0\\
& & & bcls & 96.8 & 83.3 & 75.4 & 74.3 & 82.5\\

\midrule

\multirow{4}{*}{MEDIC} & \multirow{4}{*}{\checkmark} & \multirow{4}{*}{\checkmark} & - & 95.1 & 83.7 & 73.7 & 75.5 & 82.0\\
& & & cls & 95.4 & 84.7 & 77.5 & 76.8 & 83.6\\
& & & bcls & 95.8 & \textbf{85.8} & \textbf{78.3} & \textbf{79.5} & \textbf{84.9}\\
& & & share & 96.2 & 85.2 & 77.6 & 79.4 & 84.6\\
\bottomrule
\end{tabular}
}

\vspace{-0.15in}
\label{tab: pacs6-res50-bcls}
\end{table}

\subsection{Ablation Study}
\textbf{On the effect of different learning paradigms.}
We conduct ablation studies to reveal the importance of the dualistic meta-learning scheme in our method.
To this end, we compare MEDIC with the baseline ERM \cite{naumovich1998statistical} and the most related method MLDG \cite{li2018learning}.
The core idea of MLDG is to simulate the virtual target domain at each learning iteration, regardless of the relationship among classes.
When using the same loss function and model architecture, the training strategy becomes the only variable between MEDIC and other methods.
As shown in \cref{tab: pacs6-res50-bcls}, MEDIC outperforms the above methods no matter which option is uniformly appointed, proving that dualistic gradient matching plays a critical role in improving open set performance.
Moreover, after switching from cross-entropy loss (\ie, training with close set classifier only) to open set loss function (\ie, training with the two classifiers), our method ushers in the largest performance gain on both H-score and OSCR by the same average value of $2.9\%$, indicating that the proposed strategy has better compatibility with the multi-binary classifier to learn a generalizable boundary for each known class.

\begin{figure}[tp]
    \vspace{-0.05in}
    \centering
	\begin{subfigure}{0.29\linewidth}
		\centering
        \includegraphics[width=1\linewidth]{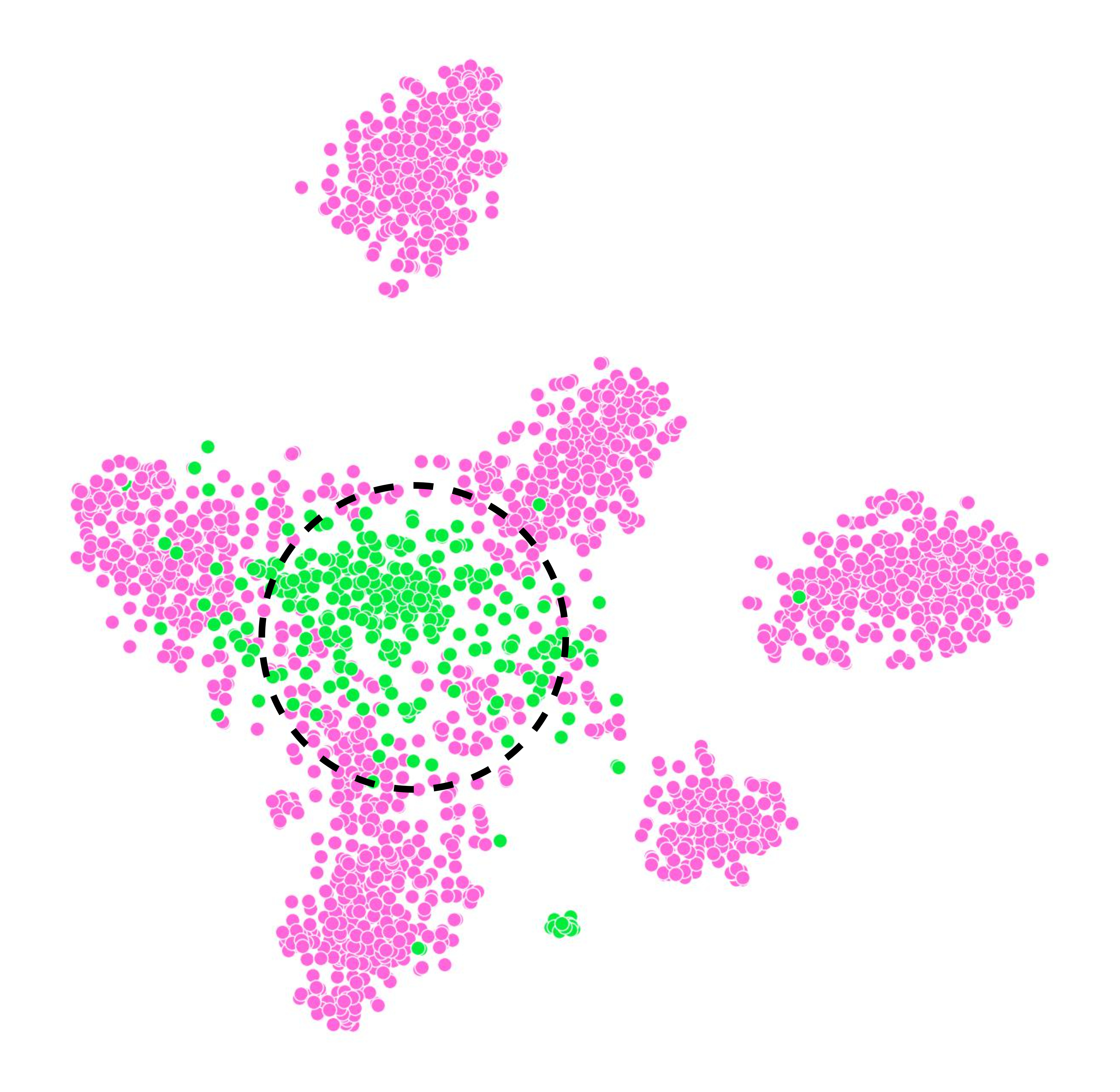}
		\caption{ERM}
	\end{subfigure}
    \quad
	\begin{subfigure}{0.29\linewidth}
		\centering
		\includegraphics[width=1\linewidth]{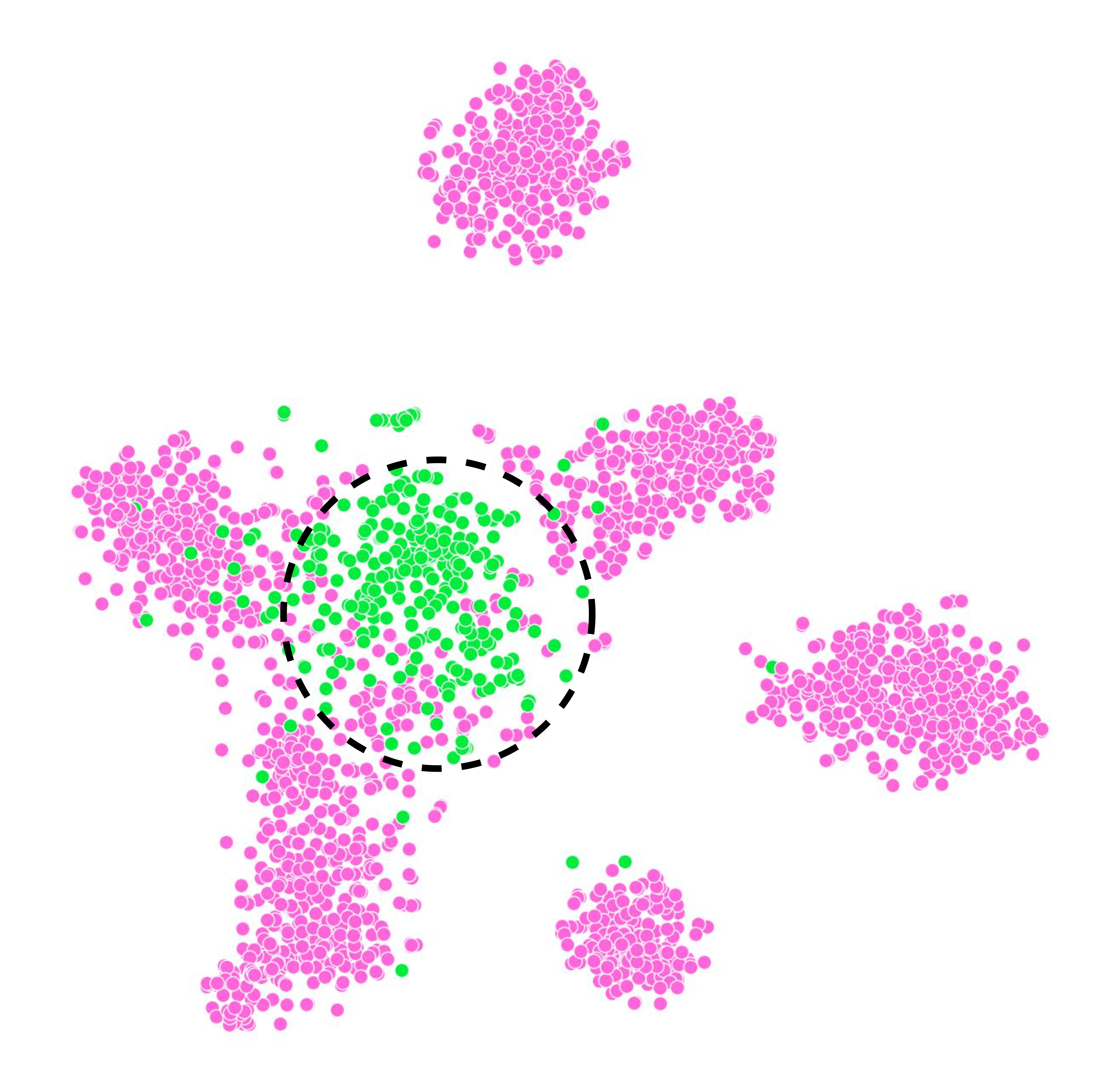}
		\caption{MLDG}
	\end{subfigure}
    \quad
    \begin{subfigure}{0.29\linewidth}
		\centering
		\includegraphics[width=1\linewidth]{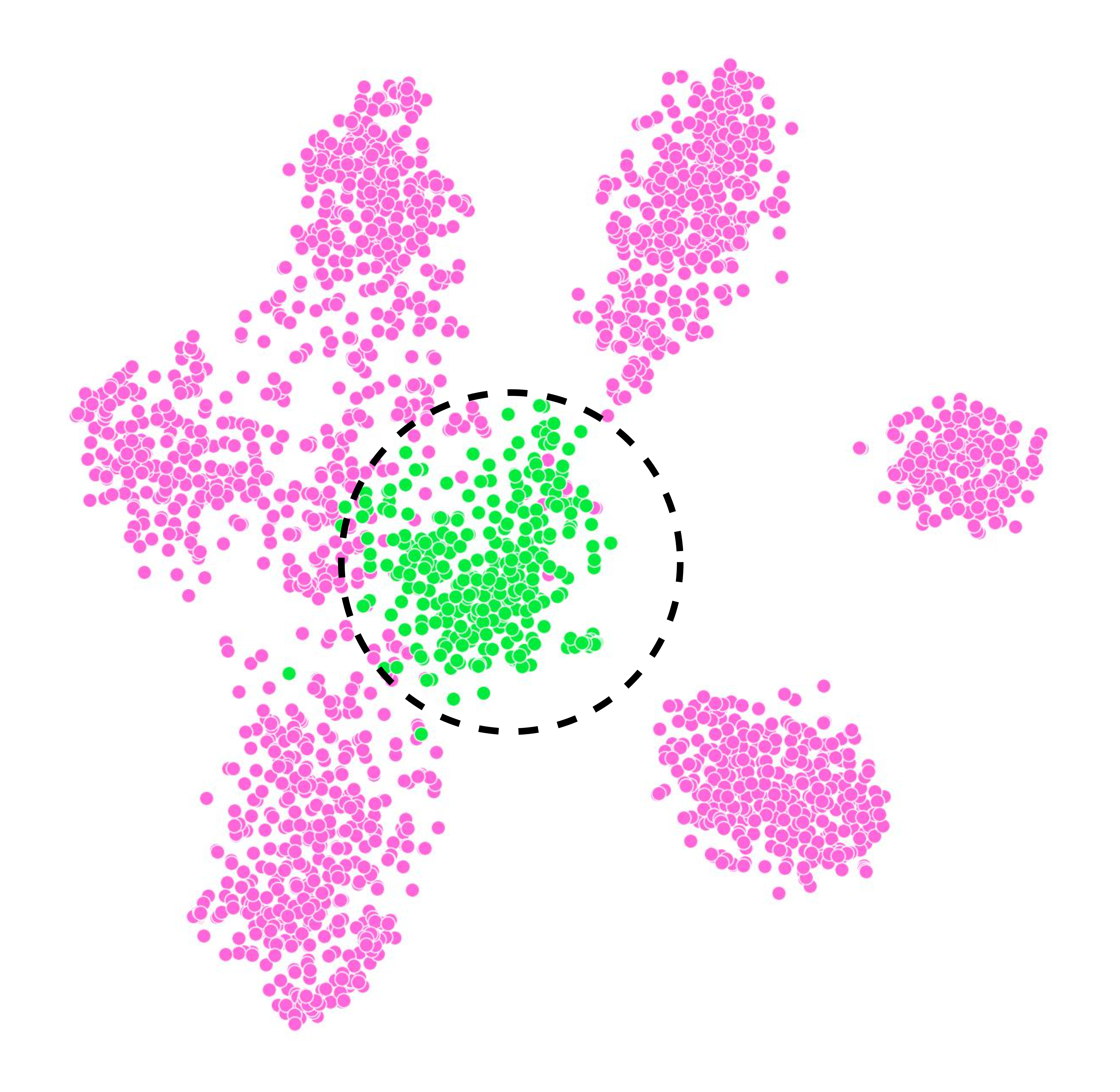}
		\caption{MEDIC}
	\end{subfigure}
    \vspace{-0.05in}
	\caption{
 T-SNE results of features in the target domain, where pink and green represents known and unknown classes respectively.
    }
    \vspace{-0.1in}
    \label{fig:t-sne}
\end{figure}

\textbf{Verify the unbiased decision boundaries.}
In practice, the shape and position of decision boundaries are influenced by the feature extractor. 
An unbiased decision boundary implies a clear separation between known and unknown classes, while a biased one indicates that the features of known classes tend to resemble those of unknown, making it difficult to differentiate between them.
We provide t-SNE results of feature representations in \cref{fig:t-sne}. 
It can be observed that unknown classes are typically distributed around the central region. In MEDIC, the overlap between known and unknown classes seems smaller, while reserving a wider space for other potential unknown classes.

\begin{figure}[tp]
    \vspace{0.05in}
    \centering
    \includegraphics[width=0.95\linewidth]{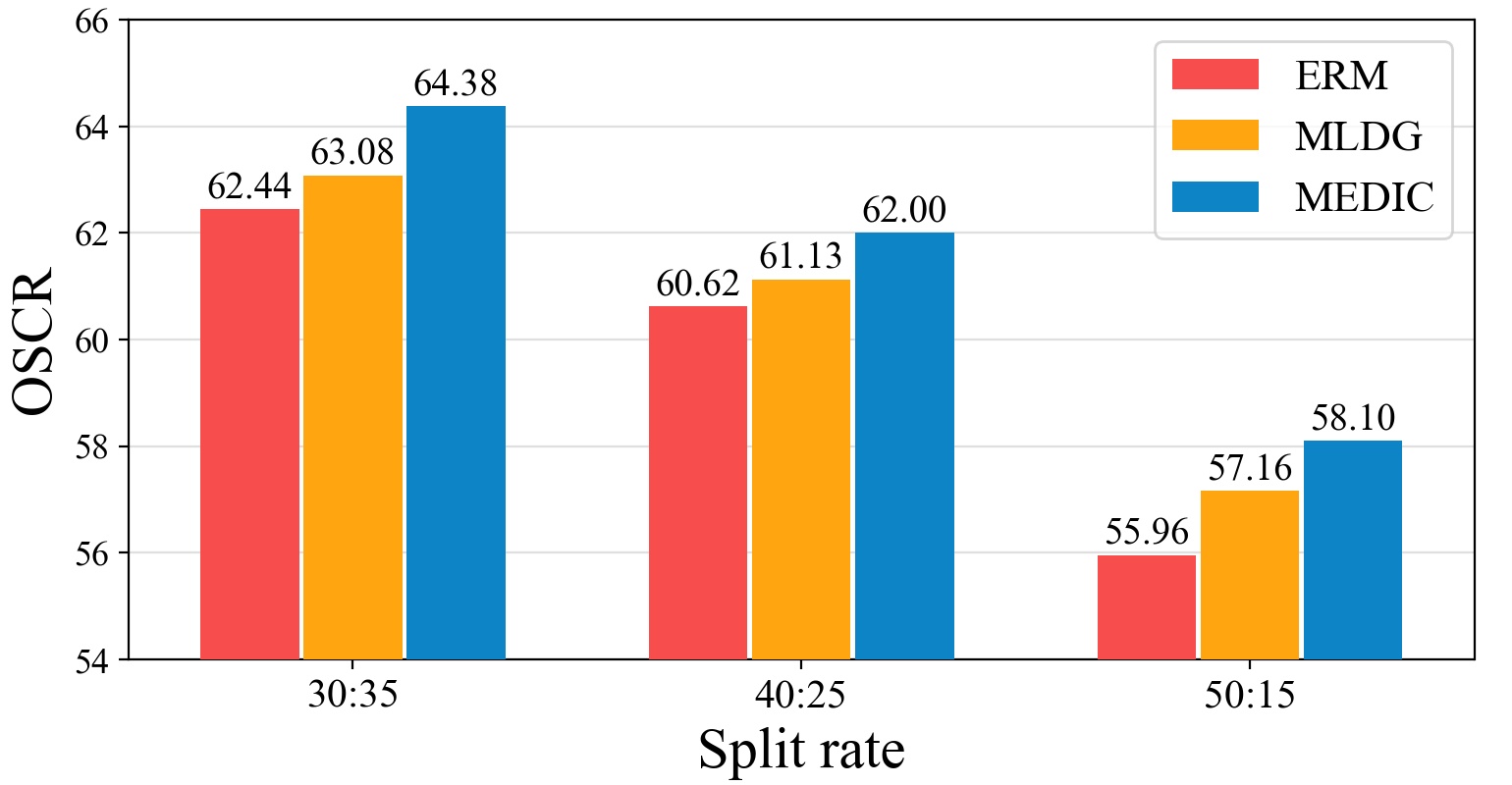}
    \vspace{-0.05in}
    \caption{The values of OSCR (\%) with the varying known-unknown divisions on Office-Home using ResNet18.}
    \vspace{-0.1in}
    \label{fig:office-range}
\end{figure}

\textbf{Varying the ratio of known to unknown classes.}
The openness reflects the proportion of unknown target classes at test time.
We conduct experiments on Office-Home to verify the adaptability of our model at different levels of openness.
The results of OSCR are visually displayed in \cref{fig:office-range}.
It is obvious that increasing the number of classes poses greater challenges to the classification task, resulting in lower accuracy of the model.
On the other hand, MEDIC continues to outperform others with noticeable improvement, indicating the robustness of our method in various scenarios.

\begin{table}[t]
\renewcommand\arraystretch{0.95}
\caption{Close set accuracy (\%) of PACS on ResNet50.}
\vspace{-0.05in}
\centering
\resizebox{0.8\linewidth}{!}{
\begin{tabular}{lccccc}
\toprule
\textbf{Method} & {\textbf{P}} & {\textbf{A}} & {\textbf{C}} & {\textbf{S}} & {\textbf{Avg}} \\
\midrule

IRM \cite{arjovsky2020invariant} & 96.7 & 84.8 & 76.4 & 76.1 & 83.5 \\
MetaReg \cite{balaji2018metareg} & \textbf{97.6} & 87.2 & 79.2 & 70.3 & 83.6 \\
ERM \cite{naumovich1998statistical} & 97.4 & 85.7 & 77.1 & 76.6 & 84.2 \\
MMD \cite{li2018domain} & 96.6 & 86.1 & 79.4 & 76.5 & 84.7 \\
MLDG \cite{li2018learning} & 97.4 & 85.5 & 80.1 & 76.6 & 84.9 \\
RSC \cite{huang2020self} & 97.6 & 85.4 & 79.7 & 78.2 & 85.2 \\
Mixstyle \cite{zhou2020domain} & 96.6 & 86.8 & 79.0 & 78.5 & 85.2 \\
CORAL \cite{sun2016deep} & 97.5 & 88.3 & 80.0 & 78.8 & 86.2 \\
DSON \cite{seo2020learning} & 96.0 & 87.0 & 80.6 & 82.9 & 86.6 \\
SWAD \cite{cha2021swad} & 97.3 & \textbf{89.3} & 83.4 & 82.5 & 88.1 \\
\midrule
MEDIC & 97.3 & 88.5 & \textbf{84.4} & \textbf{83.0} & \textbf{88.3}\\
\bottomrule
\end{tabular}
}
\vspace{-0.05in}
\label{tab: pacs7-res50}
\end{table}
\begin{table}[t]
\renewcommand\arraystretch{0.95}
\caption{Close set accuracy (\%) of Office-Home on ResNet50.}
\vspace{-0.05in}
\centering
\resizebox{0.8\linewidth}{!}{
\begin{tabular}{lccccc}
\toprule
\textbf{Method} & {\textbf{A}} & {\textbf{C}} & {\textbf{P}} & {\textbf{R}} & {\textbf{Avg}} \\
\midrule

Mixstyle \cite{zhou2020domain} & 51.1 & 53.2 & 68.2 & 69.2 & 60.4 \\
IRM \cite{arjovsky2020invariant} & 58.9 & 52.2 & 72.1 & 74.0 & 64.3 \\
RSC \cite{huang2020self} & 60.7 & 51.4 & 74.8 & 75.1 & 65.5 \\
DANN \cite{ganin2016domain} & 59.9 & 53.0 & 73.6 & 76.9 & 65.9 \\
MMD \cite{li2018domain} & 60.4 & 53.3 & 74.3 & 77.4 & 66.4 \\
MLDG \cite{li2018learning} & 61.5 & 53.2 & 75.0 & 77.5 & 66.8 \\
ERM \cite{naumovich1998statistical} & 63.1 & 51.9 & 77.2 & 78.1 & 67.6 \\
SagNet \cite{nam2021reducing} & 63.4 & 54.8 & 75.8 & 78.3 & 68.1 \\
CORAL \cite{sun2016deep} & 65.3 & 54.4 & 76.5 & 78.4 & 68.7 \\
SWAD \cite{cha2021swad} & 66.1 & 57.7 & 78.4 & \textbf{80.2} & 70.6 \\

\midrule
MEDIC & \textbf{68.1} & \textbf{58.1} & \textbf{78.7} & 79.9 & \textbf{71.2} \\
\bottomrule
\end{tabular}
}
\vspace{-0.1in}
\label{tab: office65-res50}
\end{table}

\begin{figure*}[tp]
    \centering
    \hspace{0\linewidth}
    \begin{subfigure}{0.9\linewidth}
		\centering
		\includegraphics[width=1\linewidth]{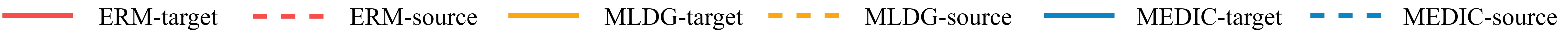}
		\label{subfig:legend-hs}
        \vspace{-0.1in}
	\end{subfigure}
    \quad
	\begin{subfigure}{0.225\linewidth}
		\centering
		\includegraphics[width=1\linewidth]{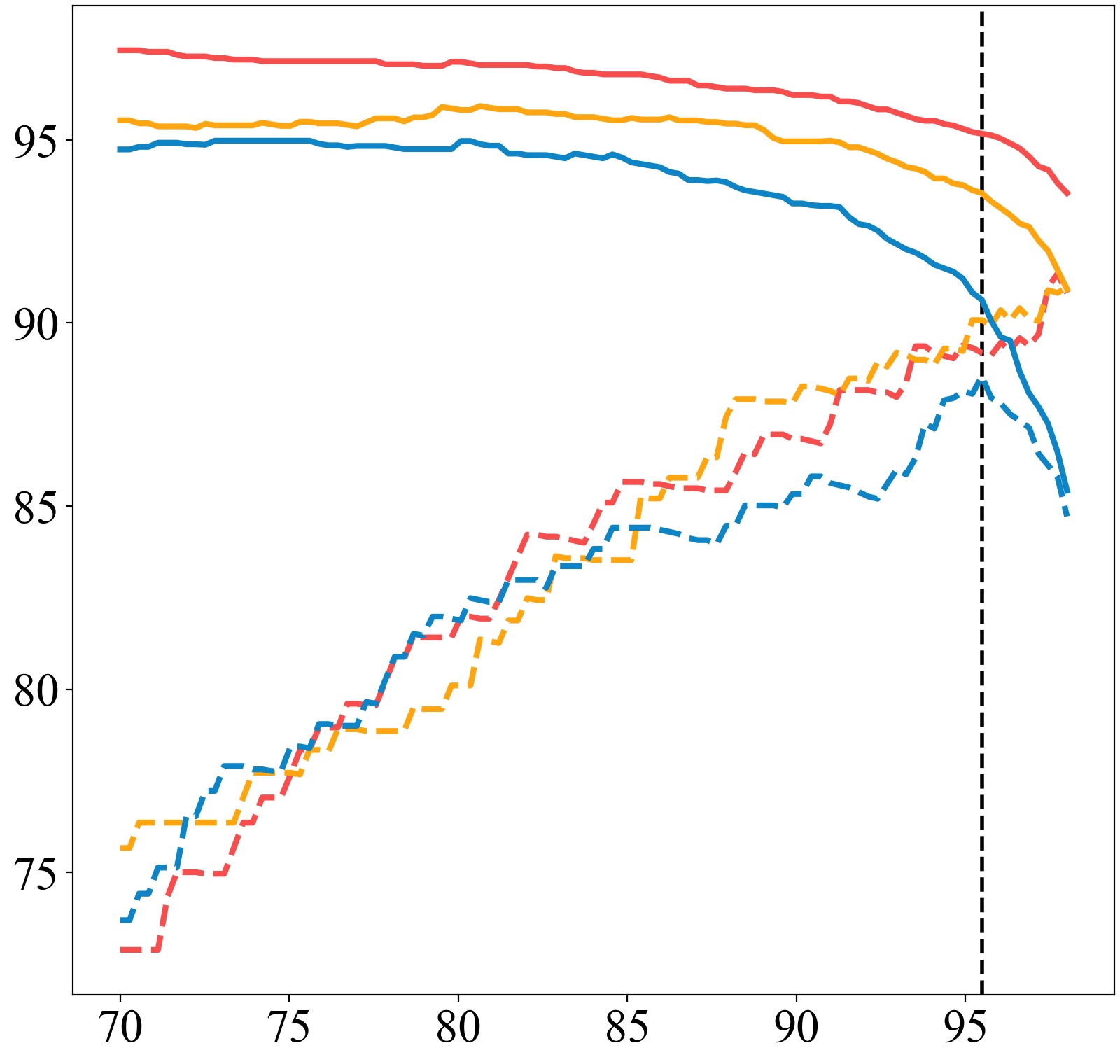}
		\caption{Photo as Target}
		\label{subfig:p-hs}
	\end{subfigure}
    \quad
	\begin{subfigure}{0.225\linewidth}
		\centering
		\includegraphics[width=1\linewidth]{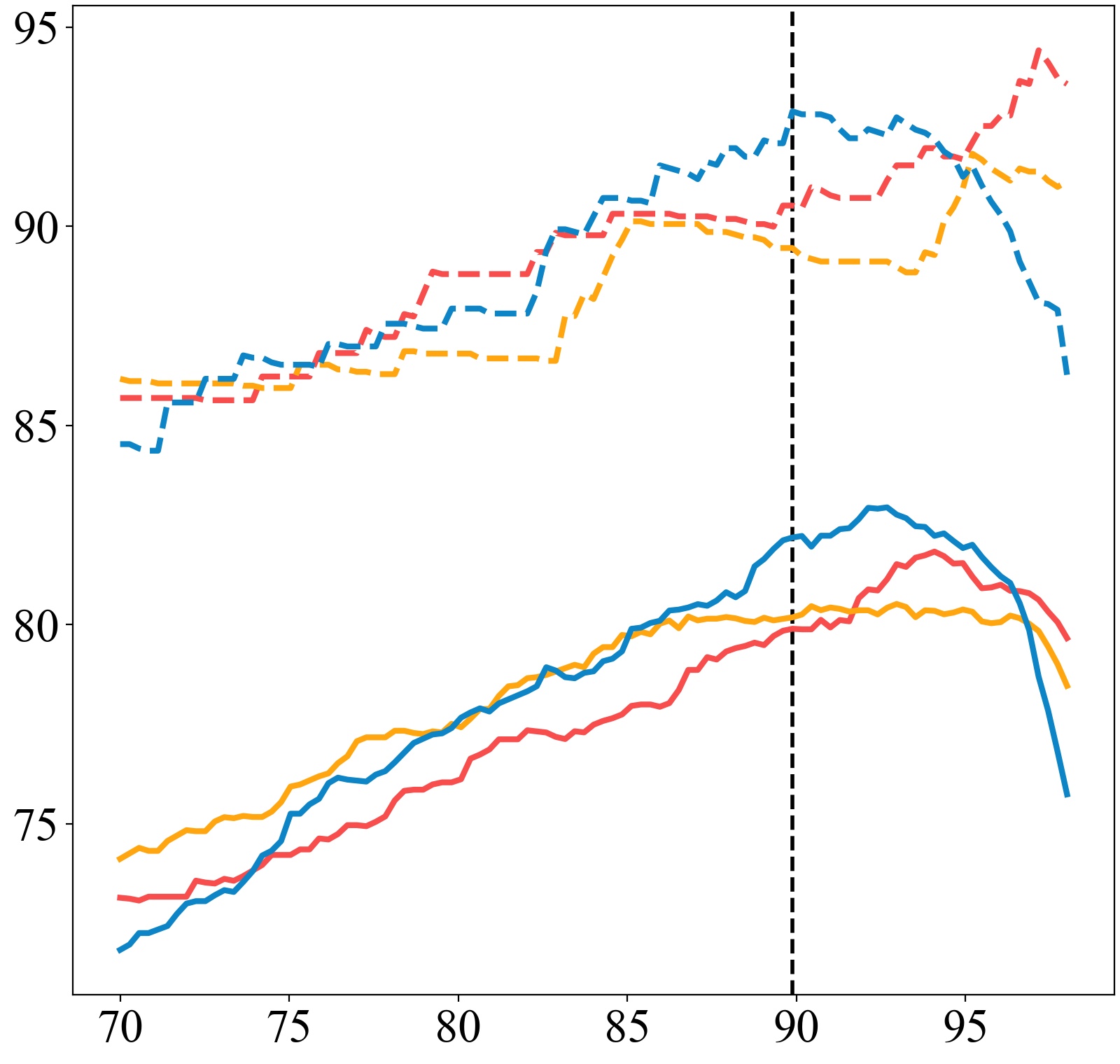}
		\caption{Art as Target}
		\label{subfig:a-hs}
	\end{subfigure}
    \quad
    \begin{subfigure}{0.225\linewidth}
		\centering
		\includegraphics[width=1\linewidth]{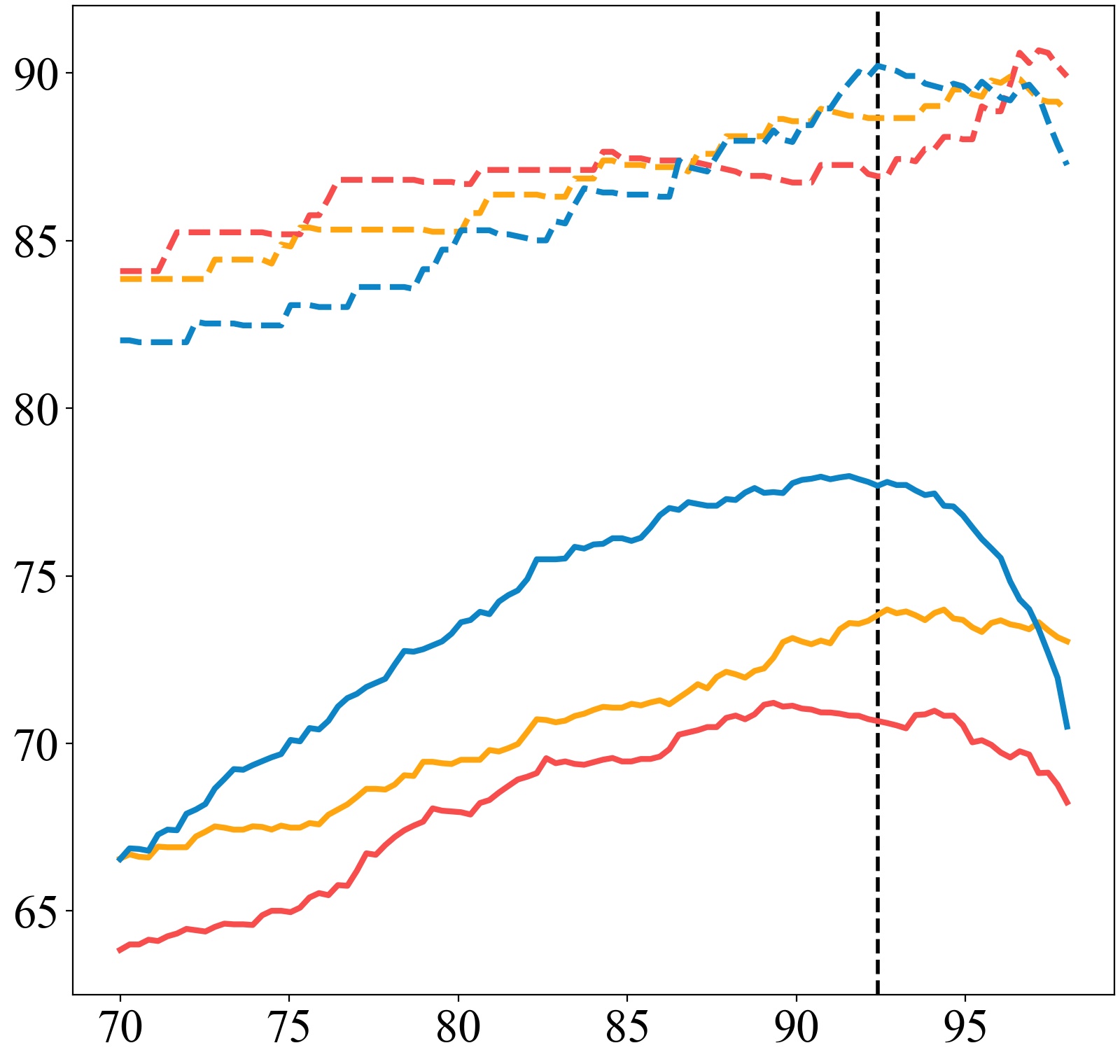}
		\caption{Cartoon as Target}
		\label{subfig:c-hs}
	\end{subfigure}
    \quad
    \begin{subfigure}{0.225\linewidth}
		\centering
		\includegraphics[width=1\linewidth]{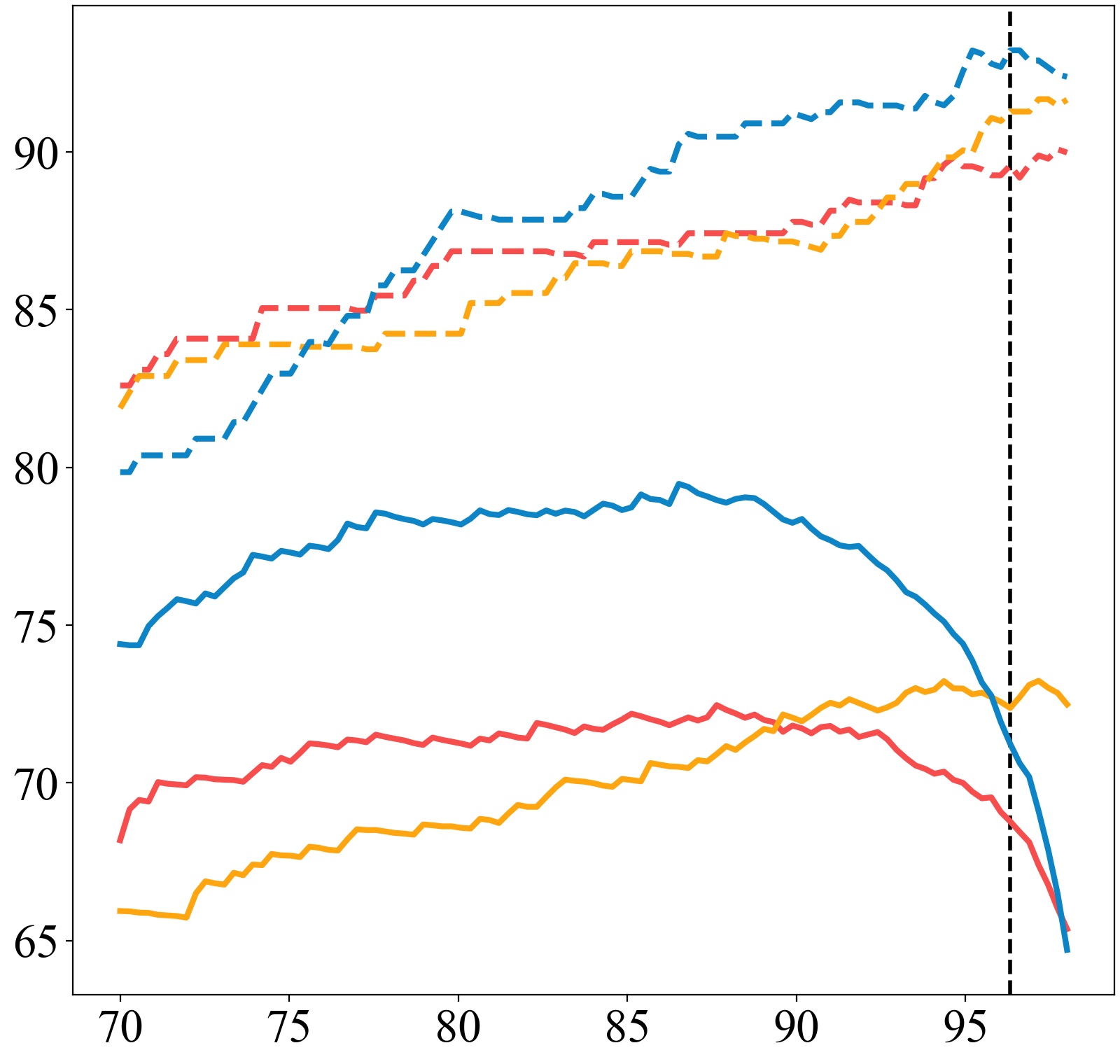}
		\caption{Sketch as Target}
		\label{subfig:s-hs}
	\end{subfigure}
    \vspace{-0.05in}
	\caption{The H-score (\%) with the varying threshold (\%) on PACS using ResNet50. 
    The solid and dashed lines represent results on the target domain and the validation set of source domains respectively. The vertical lines mark the optimal threshold on the source domains.
    }
    \vspace{-0.1in}
	\label{fig:hs-range}
\end{figure*}

\textbf{Varying the threshold of rejecting unknown classes.} 
Although OSCR provides a relatively fair way of model comparison, in practice, we have to select a specific threshold to tell knowns apart from unknowns.
The variation of H-score with thresholds on PACS is presented in \cref{fig:hs-range}.
We can observe that the proposed MEDIC consistently precedes MLDG with a large margin on two target domains: \emph{cartoon}, \emph{sketch}, and also shows a later advantage on \emph{art-painting} when the threshold is around $0.9$, illustrating that MEDIC can maintain stable performance under different thresholds.
It can be also noticed that results on \emph{photo} are usually higher than other domains and the vanilla ERM outperforms competing methods on this domain. 
This may indicate that when served as a target domain, \emph{photo} shows a higher inclination towards specific source domains, limiting the efficacy of domain-agnostic DG methods to leverage their inherent advantages.
The ideal threshold for source domains is not always reliable.
Take \emph{photo} and \emph{sketch} for example, when the model reaches its maximum H-score on the source domains, the corresponding value on the target domain is less than satisfactory, which indicates that determining the optimal threshold remains a challenging task to be addressed.

\textbf{Close set domain generalization.}
We evaluate the close set classification accuracy on PACS and Office-Home. 
As shown in \cref{tab: pacs7-res50} and \cref{tab: office65-res50}, we make a list of representative approaches in a variety of DG genres to confirm the superiority of our proposed method.
In particular, MEDIC achieves evenly matched or better performance on all target domains compared with one of the state-of-the-art algorithms SWAD \cite{cha2021swad}, which highlights the competitiveness of our method in the field of classic domain generalization.


\subsection{Analysis \& Discussion}
\textbf{Sharing parameters between classifiers.}
It can be observed that each distinct known class is associated with a unique one-vs-all classifier and a single output channel in the close set classifier. 
This channel's functioning mirrors the positive output channel of the corresponding binary classifier, implying activation by samples from the same class. 
This naturally prompts us to consider the feasibility of parameter sharing between these channels. 
In the MEDIC part of \cref{tab: pacs6-res50-bcls}, the third and fourth rows demonstrate that parameter sharing yields comparable performance to the original architecture while reducing the total output channels from $3|\mathcal{C}|$ to $2|\mathcal{C}|$. 
If we assume linearity for all classifiers, the number of reduced parameters in the improved model is equivalent to that of a complete close set classifier.

\textbf{Intuition for choosing datasets.} 
Since there are no widely acknowledged datasets for open set domain generalization, we experiment on standard domain generalization benchmarks. 
Our criteria for selecting datasets are as follows:
(i) The dataset should have been well settled in the conventional DG setting because when an unknown sample is given a low confidence score, we prefer it to be a reasonable rejection based on sufficient understanding of known classes rather than a random guess due to the lack of adequate knowledge at all.
(ii) Ideally, the semantics of each sample should correspond to an individual label. 
Suppose there are \emph{dog} and \emph{cat} in the label space and we have an image labeled as \emph{dog} that possesses the primary semantic of \emph{dog} and the secondary semantic of \emph{cat}.
In close set classification, it is reasonable to treat it as \emph{dog} because the dog holds the principal status in this image.
However, in open set experiments, some labels are converted into unknowns such as \emph{dog}.
We thus get a wrong sample with semantic of \emph{cat} but label of unknown, which can interfere with the fairness in evaluation.
Therefore, the semantics of the samples in the label space should be as pure as possible.
We hope the above two thoughts could provide a clue for future studies.

\section{Conclusion}
\label{sec:conclusion}
In this paper, we introduce the practical problem setting of open set domain generalization, which aims to tackle the challenges of domain shift and category shift concurrently in the unseen target domain.
To address this, we propose a simple yet effective meta-learning-based framework, which considers both domain-wise and class-wise gradient matching simultaneously, along with a multi-binary classifier to learn a balanced decision boundary for each known class.
We conduct experiments using comprehensive evaluation metrics on multiple benchmarks and demonstrate its superior performance in both close set and open set scenarios.

{\small
\bibliographystyle{ieee_fullname}
\bibliography{11_references}
}

\ifarxiv \clearpage \appendix
\label{sec:appendix}

 \fi

\end{document}


\title{\paperTitle \\ Supplemental Material}
\author{\authorBlock}
\maketitle

\appendix
\label{sec:appendix}


{\small
\bibliographystyle{ieee_fullname}
\bibliography{11_references}
}